\def\year{2020}\relax
\documentclass[letterpaper]{article} 
\usepackage{aaai20}  
\usepackage{times}  
\usepackage{helvet} 
\usepackage{courier}  
\usepackage[hyphens]{url}  
\usepackage{graphicx} 
\usepackage{graphicx}  
\usepackage{multirow}
\usepackage{bm}

\newcommand{\tabincell}[2]{\begin{tabular}{@{}#1@{}}#2\end{tabular}} 

\title{Dynamic Knowledge Routing Network For Target-Guided Open-Domain Conversation}
\author{Jinghui Qin\textsuperscript{\rm 1}\thanks{Equal Contribution and work was done at Dark Matter AI Inc.}, Zheng Ye\textsuperscript{\rm 1$\ast$}, Jianheng Tang\textsuperscript{\rm 1}, Xiaodan Liang\textsuperscript{\rm 1,2}\thanks{Corresponding Author.}\\
\textsuperscript{\rm 1} Sun Yat-Sen University, \textsuperscript{\rm 2} Dark Matter AI Inc.\\
{qinjingh@mail2.sysu.edu.cn, yezh7@mail2.sysu.edu.cn, sqrt3tjh@gmail.com, xdliang328@gmail.com}
}
\begin{document}

\maketitle

\begin{abstract}
Target-guided open-domain conversation aims to proactively and naturally guide a dialogue agent or human to achieve specific goals, topics or keywords during open-ended conversations. Existing methods mainly rely on single-turn data-driven learning and simple target-guided strategy without considering semantic or factual knowledge relations among candidate topics/keywords. This results in poor transition smoothness and low success rate. In this work, we adopt a structured approach that controls the intended content of system responses by introducing coarse-grained keywords, attains smooth conversation transition through turn-level supervised learning and knowledge relations between candidate keywords, and drives an conversation towards an specified target with discourse-level guiding strategy. Specially, we propose a novel dynamic knowledge routing network (DKRN) which considers semantic knowledge relations among candidate keywords for accurate next topic prediction of next discourse. With the help of more accurate keyword prediction, our keyword-augmented response retrieval module can achieve better retrieval performance and more meaningful conversations. Besides, we also propose a novel dual discourse-level target-guided strategy to guide conversations to reach their goals smoothly with higher success rate. Furthermore, to push the research boundary of target-guided open-domain conversation to match real-world scenarios better, we introduce a new large-scale Chinese target-guided open-domain conversation dataset (more than 900K conversations) crawled from Sina Weibo. Quantitative and human evaluations show our method can produce meaningful and effective target-guided conversations, significantly improving over other state-of-the-art methods by more than 20\% in success rate and more than 0.6 in average smoothness score.
\end{abstract}

\begin{table}[t]
\centering
\resizebox{.9\columnwidth}{!}{
\begin{tabular}{l}
\hline
Target: \textbf{\underline{turtle}} \\
\hline
Baseline Kernel Conversation \\
\hline
A: Hi there. How are things? \\
H: \textbf{Dressing} up, acting, skin care, and eating out... you? \\
{\tabincell{l}{ A: Not too bad. I distracted myself with my favorite things, \\ \quad \  the \textbf{color} orang, roses, etc.}} \\
H: Sounds good. Do you have a \textbf{job}? \\
A: I've an animal, a \textbf{\underline{turtle}}. Do you have any ? \\
\hline
DKRN Conversation \\
\hline
A: Hi there. How are things? \\
H: Good thanks. So, do you \textbf{read}? \\
{\tabincell{l}{ A: Not really. I like to spend time on the \textbf{farm}. Do you like \\ \quad \ sports?}} \\
H: Nah, I am not into sports. I want to produce \textbf{music}. \\
A: Cool. I also have a pet \textbf{lizard}, do you have any pets? \\
H: Nope, no \textbf{pets}. I wish i did, but one of my sisters is allergic. \\
A: That is so sad. I have a \textbf{\underline{turtle}} because mommy is allergic. \\
\hline
\end{tabular}
}
\caption{A conversation example between human (H) and agents (A). Keywords predicted by the agents or mentioned by human are highlighted in \textbf{bold}. The target achieved at the end of a conversation is \underline{underlined}.
Our method can produce more meaningful responses than \textbf{Kernel}~\cite{tang2019target} due to the semantic knowledge relation injection and dual discourse-level target-guided strategy shown in Figure~\ref{fig:model}.}
\label{tab:intro}
\end{table}

\section{Introduction}
Open-domain Conversation systems have attracted more and more attention, but creating an intelligent conversation agent that can chat with human naturally is still challenging due to the difficulty of accurate natural language understanding and generation. Many researchers have devoted themselves to overcome this challenge and achieved impressive progress from early rule-based systems~\cite{weizenbaum1983eliza} to recent end-to-end neural conversation systems that are built on massive data~\cite{vinyals2015neural,lewis2017deal} and common-sense knowledge~\cite{ghazvininejad2018knowledge,liu2018knowledge}. However, there are some limitations in current open-domain dialogue systems. First, they~\cite{yi2019towards} suffer from a generic response problem that a system response can be an answer to a large number of user utterances, such as ``I don't know.". Second, they~\cite{ram2018conversational} are struggling to interact with human engagingly. Finally, they~\cite{li2016persona} often generate inconsistent or uncontrolled responses. Moreover, in many real-world open-ended open-domain conversation application, there are some specific goals need to be achieved, such as accomplishing nursing in therapeutic conversation, making recommendation and persuasion in shopping guide, inspiring ideas in education, etc. Therefore, there is a strong demand to integrate goals and strategy into real-world open-domain conversation systems to \emph{proactively guide} agent or human to achieve specific goals during open-ended conversations. 
However, some challenges should be attacked while we integrate goals and discourse-level strategy into the open-domain conversation: 1) how to define a goal for an open-domain conversation system. 2) how to utilize a dialogue strategy in the response retrieval or generation process. 3) how to attain a general method without being tailored towards specialized goals requiring domain-specific annotations~\cite{yarats2018hierarchical,li2018towards}. 

To attack the above challenges, a very recent work ~\cite{tang2019target} make the first step to build an open-domain dialogue agent with conversational goals, which decouples the whole system into separate modules and addresses those challenges at different granularity. They enforce target-guide conversations by training a keyword predictor and a response retrieval module with turn-level supervised learning and applying a simple discourse-level target-guided rule. Nevertheless, they don't consider the semantic knowledge relations among candidate keywords which can effectively reduce the keyword selection space based on conversation context and goal, so an agent can not choose a keyword more relevant to the conversation context and goal more easily, leading to a poor smooth transition. Moreover, their simple discourse-level target-guided strategy, which chooses next keyword closer to the end target than those in preceding turns, will lead to reply repeatedly while the keyword predictor chooses the target keyword but the retrieval response does not contain the target and the human input same utterances. This behavior about repeated replies fails the conversation easily in the limited turns. 

\begin{figure*}[htbp] 
	\centerline{\includegraphics[width=0.8\textwidth]{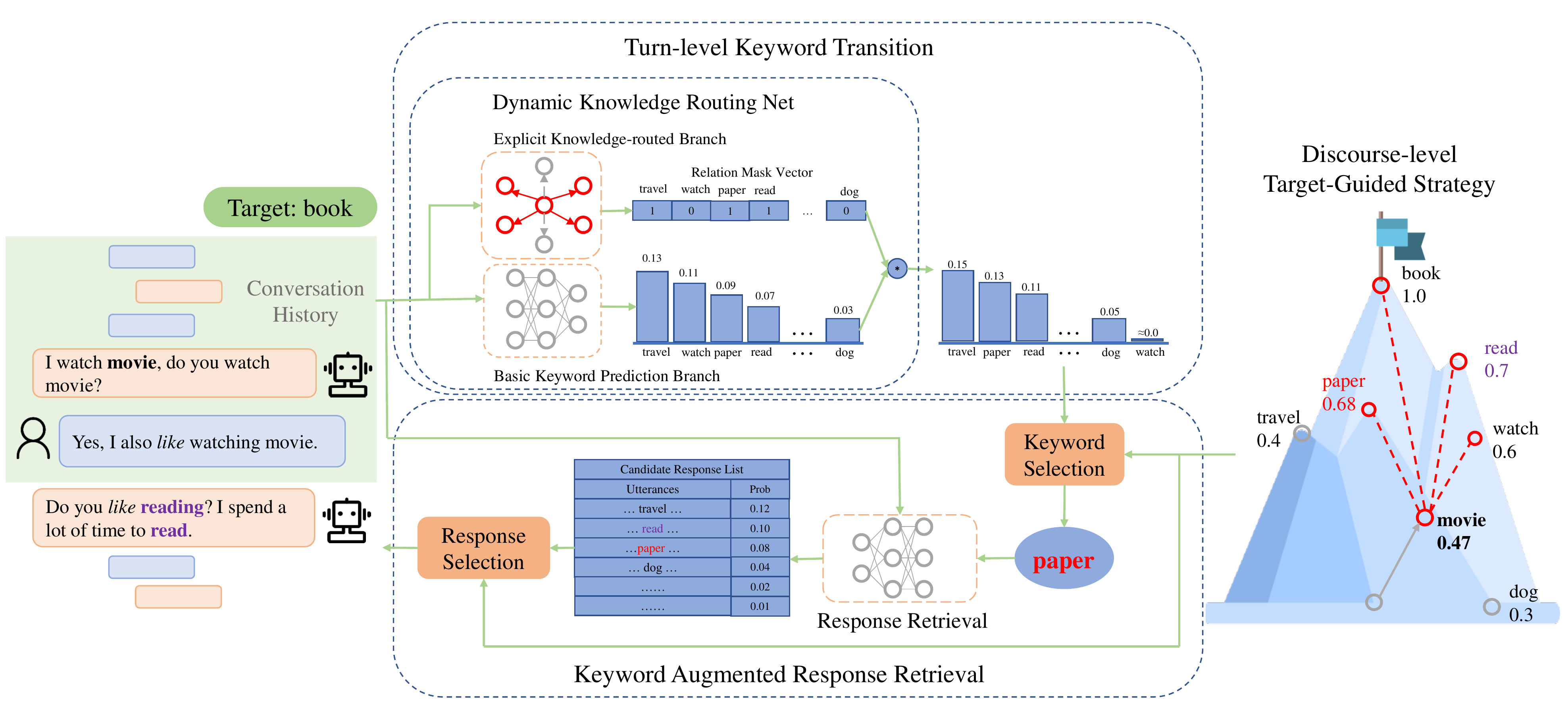}}
	\caption{An overview of our DKRN model. The left panel shows an on-going conversation with a target \emph{book}. The turn-level keyword transition module computes a distribution over candidate keywords based on conversation context and relations among candidate keywords. The discourse-level target-guided module picks a set of valid candidate keywords for the next response. The most likely valid keyword (paper) is selected and fed into the keyword-augmented response retrieval module for producing the candidate utterances of the next response. The most likely valid response contains \emph{read} is selected as the next response.}
	\label{fig:model}
\end{figure*}

To address the above problems, we propose a novel dynamic knowledge routing network (DKRN) to inject semantic knowledge relations among candidate keywords, which is built on the prior information exists in the conversation corpus, into single-turn keyword prediction for more smooth keyword transition. As shown in Figure~\ref{fig:model}, there are two branches in our DKRN: basic keyword prediction branch and explicit knowledge routing branch. When the conversation history is inputted to our DKRN, the basic keyword prediction branch will generate a rough keyword distribution. Meanwhile, our explicit knowledge routing branch refines the rough keyword distribution with the dynamic knowledge-routed vector built on the current conversation context and the target keyword. With this way, our DKRN can effectively mask all the keywords irrelevant to the current conversation context to achieve better keyword prediction. As shown in Table~\ref{tab:intro}, thanks to the semantic knowledge relation injection, our system can predict more meaningful keywords than Kernel~\cite{tang2019target}. Besides, to improve the success rate and avoid reply repeatedly, we propose a simple but effective dual discourse-level target-guided strategy. Our strategy will not only be applied to next keyword selection but also be used to choose next appropriate utterance from the results of our keyword-augmented response retrieval module to make a better trade-off between transition smoothness and the success rate of target achievement. 

Moreover, to push the research boundary of target-guided open-domain conversation to match real-world scenarios better, we build a new large-scale Chinese target-guided open-domain conversation dataset (more than 900K conversations) crawled from Sina Weibo, where a conversation consists of multiple consecutive comments under the same message. The discourse-level comments cover a broad range of topics such as shopping, disease, news, etc. The discussion topics change frequently during conversations and the data are from the real world. Therefore, these properties make our dataset particularly suitable for learning smooth, natural conversation transitions at turn-level.

Overall, our contributions can be summarized as follows.
\begin{itemize}
	\item We propose a novel dynamic knowledge routing network (DKRN) that injects the semantic knowledge relations among candidate keywords into turn-level keyword prediction for smooth keyword transition.
	\item A simple but effective dual discourse-level target-guided strategy is proposed for a better trade-off between transition smoothness and target achievement.
	\item A new challenging target-guided open-domain conversation dataset is constructed to push the research boundary of target-guided open-domain conversation to math real-world scenarios better.
	\item Extensive experiments on two target-guided open-domain conversation datasets show the superiority of our approach, which significantly surpasses state-of-the-art methods in keyword prediction accuracy, retrieval accuracy, and success rate of conversation under automatic metrics as well as human evaluation.
\end{itemize}

\section{Related Works}
Dialogue systems can be roughly categorized into two groups by applications: task-oriented dialogue systems and open-domain dialogue systems. Task-oriented dialogue system aims to achieve a specific task, e.g., movie booking~\cite{lipton2018bbq}, negotiations~\cite{cao2018emergent}, medical diagnosis\cite{xu2019end}, etc. Different from task-oriented dialogue systems, open-domain dialogue systems is designed to chat with humans on open domains without specific goals, focusing on providing reasonable responses. In recent, a lot of novel neural network architectures\cite{zhou2018multi,chen2019sequential} have been developed to improve response generation or retrieval task performance by training on the large open-domain chit-chat datasets~\cite{lowe2015ubuntu,li2017dailydialog}. Although these architectures can improve the ability of dialogue systems, there are several limitations in the current systems, such as dull responses~\cite{jiang2018sequence}, inconsistent persona~\cite{li2016persona}, etc.

Recently, ~\cite{tang2019target} proposed a new dialogue task that defines goals for open-domain chatting and creates system actions representations at the same time. In their work, they used predicted keywords as a non-parametric representation of the intended content for the next system response. Due to the lack of full supervision data, they further proposed a divide-and-conquer approach which divides the task into two competitive subtasks, called keyword prediction and response retrieval. These subtasks are conquered with target-guided rules to achieve target-guided open-domain conversation. Although their approach can achieve impressive performance for target-guided open-domain conversation, they did not consider the semantic knowledge relationship between keywords. To address the above problem, we propose a novel keyword prediction network called Dynamic Knowledge Routing Network (DKRN) which can utilize the semantic knowledge relations among candidate keywords to mask those words irrelevant to the current conversation context. With the help of the semantic knowledge relations, we can effectively mask those keywords that are uncorrelated directly to the conversation context/history, leading to effectively reduce dialogue action space under the setting of target-guided conversation and predict next keyword/subject more accurately. Thus our agent can manage dialogue states better and guide the conversation better to a designated target subject. Besides, we further introduce a new large-scale Chinese target-guided open-domain conversation dataset, which pushes the research boundary of target-guided open-domain conversation to match real-world scenarios better.

\section{Proposed Method}
\subsection{Target-guided Open-domain Conversation}
We briefly describe the definition of target-guided open-domain conversation. Given a target and a starting utterance of the conversation, an agent is asked to chat with a user starting from an arbitrary topic, and proactively guide the conversation to the target. Only the agent is informed of the target. The conversation starts with an initial topic which is randomly picked by a user. At each turn, the agent produces an utterance as a response, based on the conversation history consisting of a sequence of utterances produced by either the user or the agent. The response should satisfy two complementary and competitive objectives: 1) transition smoothness and 2) target achievement. The first objective means that the agent should produce a natural and appropriate response in the current conversation context while the second objective means the agent should proactively guide the conversation to reach the designated target. Since these two objectives are competitive, an agent needs to apply a good conversation strategy to make a trade-off between these two different objectives. Following  ~\cite{tang2019target}, we define a target to be a keyword, e.g., an named entity Costco, or a common noun football, etc. We consider a target is reached when an human or agent utterance contains the target or similar word.

\subsection{Method Overview}
Since there is not any large-scale public dataset suitable for end-to-end learning discourse-level conversation strategy in open domain under the target-guided problem setting, we adopt a divide-and-conquer approach to attack the two complementary and competitive objectives separately. Specially, for transition smoothness, we propose dynamic knowledge routing network (DKRN) and adopt keyword-augmented response retrieval module~\cite{tang2019target} to maintain smooth conversation transition at turn-level by turn-level supervised learning on open-domain chat data. Our DKRN will utilize the relation graph between candidate keywords which is built on open-domain chat data to mask irrelevant candidate keywords in term of current conversation context, leading to effectively reduce the coarse-grained keyword search space. Our empirical results demonstrate that this way can effectively improve the accuracy of turn-level keyword prediction and response retrieval. Turn-level accuracy is crucial for smooth discourse-level conversation. Then, we apply a simple but effective dual rule-based discourse-level target-guided strategy to ensure the agent can guide the conversation to achieve the target with enough turns of dialogue. To make a good trade-off between transition smoothness and target achievement, we decouple the decision-making process and utterance generation by injecting target-guided strategy which explicitly models the intended coarse-grained keywords in the next agent utterance. Thus, our structured solution consists of several core modules, including a dynamic knowledge routing network (DKRN) for turn-level keyword transition prediction, a dual rule-based discourse-level target-guided strategy for discourse-level target guide, and a keyword-augmented response retrieval module for next response retrieval. 

\subsection{Dynamic Knowledge Routing Network (DKRN)}
Given the conversation history at each dialogue turn, DKRN aims to predict an appropriate keyword for the next response in terms of the current conversation context. Since DKRN is agnostic to the end target, we can adopt any open-ended chitchat data with extracted utterance keywords to train DKRN in a supervised manner with the conventional objective. Adopting a neural network as a backbone to end-to-end predict the next keyword is a common practice in recent conversation systems~\cite{zhou2018multi,tang2019target}. However, they didn't consider the semantic knowledge relations among candidate keywords for keyword prediction. To utilize the explicit relation knowledge among candidate keywords, we propose a novel Dynamic Knowledge Routing Network (DKRN) by considering prior relations between keywords to predict more reasonable keyword for the next response. As shown in Figure~\ref{fig:model}, our DKRN consists of two branches: basic keyword prediction branch and explicit knowledge-routed branch. 

\textbf{Basic Keyword Prediction Branch.} We first utilize a basic keyword prediction branch to generate a rough keyword distribution. More concretely, we first use a recurrent neural network to encode the current conversation context $h_t$ and feed the resulting feature to two continuous fully connected layers to obtain a distribution over keywords $\bm{K}_{t+1}$ for the next turn t+1. The activation function of the first fully connected layer is rectified linear unit. Thus, we can formulate our basic keyword prediction branch as follows:
\begin{equation}
	\bm{K}_{t+1} = fc_2(\sigma(fc_1(RNN(h_{t}))))
\end{equation}
where $\sigma(\cdot)$ is a ReLU. $\bm{K}_{t+1}$ and $h_t$ represent the predicted distribution over keywords for turn t+1 and current conversation context at turn t, respectively.

\textbf{Explicit Knowledge-routed Branch.} The keywords transition should be smooth in a good conversation. Filtering out keywords irrelevant to the current conversation context when the system predicts the next keyword is crucial for ensuring conversation smoothness. Thus, we design an explicit knowledge-routed module to generate an appropriate keyword mask to ensure turn-level keyword transition smoothness. The explicit keyword relation graph is built on the dataset. For each turn of conversation in the dataset, we first extract the keywords from the current utterance and the next utterance. Then we build the relationships between the keywords of current utterance and the keywords of next utterance as directed knowledge-routed graph. Edge only exists between two keywords where a keyword occurs in the first utterance of a dialogue turn while another keyword occurs in the second utterance. During training and inference, our explicit knowledge-routed branch works as follows: First, it will extract keyword set $S_t$ from current conversation context $h_t$. Then, it looks up all keywords $R_t$ relevant to $S_t$ from knowledge relation graph $\mathcal{G(V,E)}$, which is built on the prior information exists in the conversation corpus, to construct knowledge relation subgraph instantaneously. Finally, it will generate a mask vector $\bm{M}_{t+1}$ based on the knowledge relation subgraph. The length of $\bm{M}_{t+1}$ is equal to the number of candidate keywords and each index value is the same as the candidate keyword's id. In the mask vector, only the positions belong to the keywords in $R_t$ will be set to 1 and other positions will be set to $-1e8$. Therefore, given keyword relation graph $\mathcal{G(V,E)}$ where $\mathcal{V}$ is the set of all candidate keywords and $\mathcal{E}$ is all relations between keywords. Let $\bm{M}_{t+1}$ be the mask vector for refining the keyword prediction of next turn t+1, and suppose that there are m keywords in $S_t$ and n keywords in $R_t$. So we can formulate our explicit knowledge-routed branch as follows:
\begin{equation}
\bm{M}_{t+1}(R_{t,j}) = \left\{ \begin{array}{l}
1,\; e(S_{t,i},R_{t,j})\in \mathcal{E};\\
-1e8,\;otherwise;\\
\end{array} \right.\\
\end{equation}
where $i=1,2,..,m$ and $j=1,2,...,n$. $e(a,b)$ represents the relation between keyword $a$ and keyword $b$.

With the rough keyword distribution vector $\bm{K}_{t+1}$ and dynamic knowledge-routed vector $\bm{M}_{t+1}$, we enforce knowledge routing by redistributing the keyword distribution $\bm{K}_{t+1}$ as $\bm{P}_{t+1}$  for next keyword prediction as follows:
\begin{equation}
	\bm{P}_{t+1} = \sigma(\bm{K}_{t+1} - 1 + \bm{M}_{t+1})
\end{equation}
where $\sigma(\cdot)$ is $sigmoid$ activation function. Our DKRN is learned by maximizing the likelihood of observed keywords in the data. 

\subsection{Dual Discourse-level Target-Guided Strategy}
This module aims to automatically guide the discussing topic to reach the end target in the course of conversation. We deploy a dual discourse-level target-guided strategy to make a better trade-off between transition smoothness and target achievement than the strategy~\cite{tang2019target} under the limited dialogue turns. In target-guided open-domain conversation, the agent should be constrained to predict a keyword strictly closer to the end target at each turn, compared to those in preceding turns. Different from the existing strategy, our strategy not only constrains the keyword predicted for keyword-augmented response retrieval must be strictly closer to the end target but also constrains the next chosen response must contain the predicted keyword or a keyword closer to the end target than the previous predicted keyword. Figure~\ref{fig:model} illustrates the dual strategy. Given the keyword \emph{movie} of the current turn and its closeness score (0.47) to the target \emph{book},  only those keywords with higher target closeness are the valid candidate keywords for the next turn, such as \emph{watch} (0.6), \emph{paper} (0.68), \emph{read} (0.7). In particular, we use cosine similarity between normalized word embeddings as the measure of keyword closeness. At each turn for predicting the next keyword, the above constraint first collects a set of valid candidates, and the system samples or picks the most likely one from the set according to the keyword distribution. After we get the sorted candidate response list produced by keyword-augmented response retrieval module, we apply the above constraint again and choose the first utterance which contains either the predicted keyword or other keywords closer to the end target than the predicted keyword. In this way, the next response can be both a smooth transition and an effective step towards the end target.

\subsection{Keyword-augmented Response Retrieval}
Keyword-augmented response retrieval module aims to produce an appropriate response based on both the conversation history and the predicted keyword. Inspired by ~\cite{tang2019target}, we adapt the sequential matching network (SMN) ~\cite{wu2017sequential} as the basic backbone of our keyword-augmented response retrieval module and enhance it with the predicted keyword. The response retrieval process is as follows: We first apply multiple recurrent networks to encode the input conversation history, the predicted keyword for the next response, and all candidate responses from a corpus, respectively. Then, two vectors can be obtained by computing the Hadamard product of the candidate response feature and the history feature as well as the Hadamard product of the candidate response feature and the keyword feature. Finally, we concatenate these two vectors as a new vector and feed it into a dense layer with sigmoid activation to get the matching probability of each candidate response. Based on the matching probability of each candidate response and our dual discourse-level target-guided strategy, we can select an appropriate utterance as the next response. Similar to DKRN, we can use any open-domain chitchat data to train the retrieval module in an end-to-end supervised manner. The objective is maximizing the likelihood of observed response based on the historical context and predicted keyword.

\section{Experiments}

\subsection{Weibo Conversation Dataset}
Existed dataset~\cite{tang2019target} is inefficient for agents to learn dynamic keyword transition due to its limited size and the crowdsourcing setting of the ConvAI2 corpus. To push the research boundary of the task to match real-world scenarios better, we construct a new Weibo Conversation Dataset for target-guide open-domain conversation, denoted as CWC. Our dataset is derived from a public multi-turn conversation corpus\footnote{http://tcci.ccf.org.cn/conference/2018/dldoc/trainingdata05.zip} crawled from Sina Weibo, which is one of the most popular social platforms of China. The dataset covers rich real-world topics in our daily life, such as shopping, disease, news, etc. Although the conversations are not necessarily suitable for learning discourse-level strategies, as they were originally crawled without end targets and did not exist target-guided behaviors, these properties make our dataset particularly suitable for learning smooth, natural conversation transitions at turn-level.

To make our corpus suitable for turn-level keyword transition, we augment the data by automatically extracting keywords of each utterance. Specially, we apply a rule-based keyword extractor to select candidate keywords in terms of their word frequency, word length, and Part-Of-Speech features. The keywords must meet the following conditions: 1) its word frequency is more than 2000 in the corpus; 2) its word length is more than 2; 3) its Part-Of-Speech should be noun, verb, or adjective. We split our data set randomly into three parts: train set (90\%), validation set (5\%), and test set (5\%). The data statistics of our proposed dataset is listed in Table~\ref{tab:ds}. The resulting dataset is used in our approach for training both the keyword prediction module and the response retrieval module. For training the response retrieval module, we follow the retrieval-based chit-chat literature~\cite{wu2017sequential} and randomly sample 19 negative responses for each turn as the negative samples.

\begin{table}[htbp]
\centering
\resizebox{.9\columnwidth}{!}{
\begin{tabular}{c|c|c|c|c|c|c}
\hline
\multirow{2}{*}{ }& \multicolumn{3}{c|}{\tabincell{c}{ Chinese Weibo \\ Conversation Dataset}} & \multicolumn{3}{c}{\tabincell{c}{ Target-guide \\ PersonaChat Dataset}} \\\cline{2-7}
& Train & Val & Test & Train & Val & Test\\ \hline
\#Conversations & 824,742 &45,824 &45,763 & 8,939 &500 &500 \\
\#Utterances & 1,104,438 &60,893 &60,893 & 101,935 &5,602 &5,317 \\
\#Keyword types & 1,760 &1,760 &1760 & 2,678 &2,080 &1,571\\
\#Avg. keywords & 2.6 & 2.6 & 2.6& 2.1 &2.1 &1.9 \\ 
\hline
\end{tabular}}
\caption{Data Statistics of our Chinese Weibo Conversation Dataset (CWC) and Target-guide PersonaChat Dataset (TGPC)~\cite{tang2019target}. The last row is the average number of keywords in each utterance. The vocabulary sizes of these two datasets are around 67K and 19K, respectively.}
\label{tab:ds}
\end{table}

\begin{table*}[t]
\centering
\resizebox{0.8\textwidth}{!}{
\begin{tabular}{c|c|cccc|cccc}
\hline
\multirow{2}{*}{Dataset} &\multirow{2}{*}{System} & \multicolumn{4}{c|}{Keyword Prediction} & \multicolumn{4}{c}{Response Retrieval} \\
& & $R_{w}@$1 & $R_{w}@$3 & $R_{w}@$5 & P$@$1 & $R_{20}@$1 & $R_{20}@$3 & $R_{20}@$5 & MRR \\ \hline
\multirow{5}{*}{TGPC}& Retrieval~\cite{wu2017sequential} & - & - & - & - & 0.5063	&0.7615	&0.8676	&0.6589 \\
& PMI~\cite{tang2019target} & 0.0585 & 0.1351 &0.1872 &0.0871 & 0.5441 & 0.7839 & 0.8716 & 0.6847 \\
& Neural~\cite{tang2019target} & 0.0708	&0.1438	&0.1820	&0.1321	&0.5311	&0.7905	&0.8800	&0.6822 \\
& Kernel~\cite{tang2019target} & 0.0632	&0.1377	&0.1798	&0.1172	&0.5386	&0.8012	&0.8924	&0.6877 \\
& DKRN (ours) & \textbf{0.0909} & \textbf{0.1903} & \textbf{0.2477} & \textbf{0.1685} & \textbf{0.5729} & \textbf{0.8132} & \textbf{0.8966} & \textbf{0.7110} \\
\hline
\multirow{5}{*}{CWC}& Retrieval~\cite{wu2017sequential} & - & - & - & - & 0.5785	&0.8101	&0.8999	&0.7141 \\
& PMI~\cite{tang2019target} & 0.0555 & 0.1001 & 0.1212 & 0.0969 & 0.5945 & 0.8185 & 0.9054 & 0.7257 \\
& Neural~\cite{tang2019target} & 0.0654	&0.1194	&0.1450	&0.1141	&0.6044	&0.8233	&0.9085	&0.7326 \\
& Kernel~\cite{tang2019target} & 0.0592 & 0.1113 & 0.1337 & 0.1011 & 0.6017 & 0.8234 & 0.9087 & 0.7320 \\
& DKRN (ours) & \textbf{0.0680}	& \textbf{0.1254} &\textbf{0.1548} &\textbf{0.1185} & \textbf{0.6324} & \textbf{0.8416} & \textbf{0.9183} & \textbf{0.7533} \\
\hline
\end{tabular}}
\caption{Results of Turn-level Evaluation.}
\label{tab:tle}
\end{table*}

\begin{table}[t]
\centering
\resizebox{0.9\columnwidth}{!}{
\begin{tabular}{c|cc|cc}
\hline
\multirow{2}{*}{System} & \multicolumn{2}{c|}{TGPC} & \multicolumn{2}{c}{CWC} \\\cline{2-5}
& Succ. (\%) & \#Turns  & Succ. (\%) & \#Turns  \\ 
\hline
Retrieval~\cite{wu2017sequential} & 7.16 & 4.17 & 0 & - \\
Retrieval-Stgy~\cite{tang2019target} & 47.80 & 6.7 & 44.6 & 7.42 \\
PMI~\cite{tang2019target} & 35.36 & 6.38 & 47.4 & 5.29 \\
Neural~\cite{tang2019target} & 54.76 & 4.73 & 47.6 & 5.16 \\
Kernel~\cite{tang2019target} & 62.56 & 4.65 & 53.2  &  4.08 \\
DKRN (ours) & \textbf{89.0} & 5.02 & \textbf{84.4} & 4.20 \\
\hline
\end{tabular}}
\caption{Results of Self-Play Evaluation.}
\label{tab:spe}
\end{table}

\begin{table}[t]
\centering
\resizebox{0.9\columnwidth}{!}{
\begin{tabular}{c|cc}
\hline
System & Succ. (\%) & Smoothness  \\ 
\hline
Retrieval-Stgy~\cite{tang2019target} & 54.0 & 2.48  \\
PMI~\cite{tang2019target} & 46.0 & 2.56  \\
Neural~\cite{tang2019target} & 36.0 & 2.50  \\
Kernel~\cite{tang2019target} & 58.0 & 2.48 \\
DKRN (ours) & \textbf{88.0} & \textbf{3.22} \\
\hline
\end{tabular}}
\caption{Results of the Human Rating on CWC.}
\label{tab:hr1}
\end{table}

\begin{table}[ht]
\centering
\resizebox{0.9\columnwidth}{!}{
\begin{tabular}{c|ccc}
\hline
 & \tabincell{c}{Ours \\ Better(\%)} & \tabincell{c}{No \\ Prefer(\%)} & \tabincell{c}{Ours \\ Worse(\%)} \\
\hline
Retrieval-Stgy~\cite{tang2019target} & \textbf{62} & 22 & 16 \\
PMI~\cite{tang2019target} & \textbf{54} & 32 & 14  \\
Neural~\cite{tang2019target} & \textbf{60} & 22 & 18  \\
Kernel~\cite{tang2019target} & \textbf{62} & 26 & 12  \\
\hline
\end{tabular}}
\caption{Results of the Human Rating on CWC.}
\label{tab:hr2}
\end{table}

\subsection{Experimental Setup and Training Details}
We conduct our experiments on the dataset proposed by~\cite{tang2019target}, denoted as TGPC, and our proposed Chinese Weibo Conversation Dataset, denoted as CWC. Table~\ref{tab:ds} illustrates the detailed data statistics of the two datasets. The TGPC contains more than 9K conversations and 100K utterances, covering more than 2000 keywords While the CWC contains more than 900K conversation covering 1760 keywords. We compare our system with a diverse set of state-of-the-art systems, including \textbf{Retrieval}~\cite{wu2017sequential}, \textbf{Retrieval-Stgy}~\cite{tang2019target}, \textbf{PMI}~\cite{tang2019target}, \textbf{Neural}~\cite{tang2019target}, and \textbf{Kernel}~\cite{tang2019target}, with turn-level evaluation, automatic simulation-based evaluation, and human evaluation.

For TGPC and CWC, we both apply a single-layer GRU~\cite{chung2014empirical} in our encoder. For TGPC, both the word embedding and hidden dimensions are set to 200. GloVe\footnote{https://nlp.stanford.edu/projects/glove/} is used to initialize word embeddings. For CWC, we set the word embedding and hidden dimensions as 300. Baidu Encyclopedia Word2Vec~\cite{li2018analogical} is used to initialize word embeddings. The other hyper-parameters for both datasets are the same. We apply ADAM optimization~\cite{kingma2015adam} with an initial learning rate of 0.001 and decay to 0.0001 in 10 epochs. Our approach is implemented based on Texar~\cite{hu2019texar}.

\subsection{Turn-level Evaluation}
A good conversation should be coherent and engaging, so the performance of turn-level conversation is crucial for coherent and engaging conversations. Thus, we first evaluate the performance of each conversation turn in terms of the accuracy of  turn-level keyword prediction and response retrieval. We disable our discourse-level target-guided strategy and only measure how accurate our system can predict the next keyword and select the correct response on the test data of two conversation datasets. The turn-level evaluation mainly follows the evaluation methodology adopted in ~\cite{wu2017sequential} which can effectively validate the performance of single-turn keyword-augmented conversation production.

\subsubsection{Evaluation metrics}
We evaluate the performance with two metrics for the keyword prediction task: 1) $\bm{R_w@K}$: keywords recall at position K (=1,3,5) in all candidate keywords, 2) $\bm{P@1}$: the ratios of the expected keyword at the first position in all candidate keywords. For the response retrieval task, 19 negative responses are sampled randomly for each test case, and we calculate $\bm{R_{20}@K}$ meaning recall at position K in the 20 candidate responses. Besides, we also calculate $\bm{MRR}$, the mean reciprocal rank~\cite{manning2010introduction}, which is an important metric to measure the quality of retrieval.

\begin{figure*}[t] 
	\centerline{\includegraphics[width=0.8\textwidth]{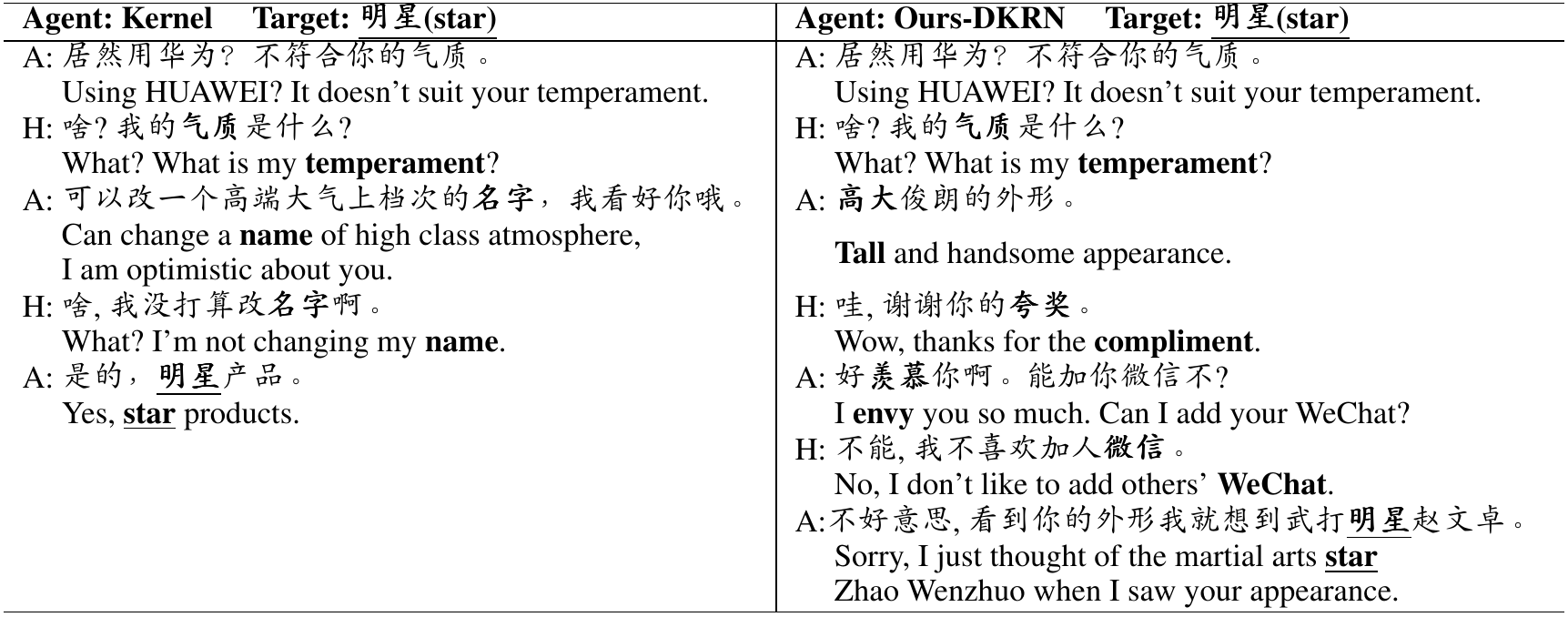}}
	\caption{A conversation example between human (H) and agents (A) with the same target and starting utterance in CWC dataset.Keywords predicted by the agents or mentioned by human are highlighted in \textbf{bold}. The target achieved at the end of a conversation is \underline{underlined}.}
	\label{fig:qs}
\end{figure*}

\subsubsection{Results}
Table~\ref{tab:tle} shows the turn-level evaluation results. Our approach outperforms all state-of-the-art methods in terms of all metrics on both datasets with two tasks. For the TGPC dataset, our DKRN surpasses other methods with a clear margin on the prediction accuracy. Moreover, our DKRN can also boost the performance of keyword-augmented response retrieval module. Even though the CWC dataset is more diverse and difficult than the TGPC dataset, our DKRN still outperforms other methods. Although the performance improvement on keyword prediction is slight, our DKRN still can boost a great performance improvement on keyword-augmented response retrieval. This indicates that our approach can maintain smooth conversation transition more effectively on turn-level.

\subsection{Target-guided Conversation Evaluation}
In this section, we conduct an automatic simulation-based evaluation and human evaluation to show our DKRN with dual discourse-level target-guided strategy is more effective and robust in the target-guided conversation setup.

\subsubsection{Self-Play Simulation}
Following prior work~\cite{lewis2017deal}, we deployed a task simulator to automatically conduct target-guided conversations. Specially, we use the Retrieval agent~\cite{wu2017sequential} as the role of human to retrieve a response without knowing the end target. At the beginning of each self-play simulation, the simulator picks a candidate keyword and an utterance as the end target and starting point randomly. Then each agent chats with its corresponding Retrieval agent and proactively guides the conversation to the given target. For TGPC, we applied WordNet\footnote{https://wordnet.princeton.edu/} to evaluate whether the target is achieved by identifying keywords semantically close to the end target. For CWC, we just simple judge whether the current conversation context contains the end target. We set a maximum number of turns as 8 to prevent from an endless conversation that can not reach the target. We use the success rate of reaching the targets (\textbf{Succ.}) and the average number of turns taken to achieve the target (\textbf{\#Turns}) as our evaluation criterion. Table~\ref{tab:spe} shows the results of 500 simulations for each of the comparison systems. Although the average number of turns of our approach is slightly more than \textbf{Kernel}, our system obtains the highest success rate, significantly improving over other approaches.

\subsubsection{User Study}
We conduct user study for a more thorough evaluation on our CWC dataset. We use the DialCrowd toolkit ~\cite{lee2018dialcrowd} to set up user study interfaces and conduct two kinds of user studies as below. The first study aims to measure system performance in terms of two key objectives, namely transition smoothness and target achievement. We construct 50 test cases, each of which has a target and a starting utterance. In each case, a user will chat with a randomly selected agent. If the agent thinks the conversation has reached the given target or the maximum number of conversation turns, the agent will inform and show the target to the user. The DialCrowd will ask the user to judge whether the target is reached and rate the transition smoothness of the conversation with a smoothness score. The smoothness score ranges from 1 (very bad) to 5 (very good). All agents are evaluated on all 50 test cases. Table~\ref{tab:hr1} shows the evaluation results. Our DKRN agent outperforms all other agents with a large margin. Besides, the success rate of our agent is consistent with itself while other agents are slightly worse than those in self-play simulation (Table~\ref{tab:spe}). 

The second study compares our agent with other agents side-by-side on 50 test cases. In each test case, a user is asked to chat with our agent and one of the other agents picked randomly. To avoid any bias, we first randomly pick one of the agent pair to chat with the user and then another. After both two conversations are ended, the user is asked to rank the two conversations. If the two conversations are equally good or bad, the user can choose ``no preference". Table~\ref{tab:hr2} shows the results of the second study. Our agent outperforms the comparison agents with a large margin. For example, compared to \textbf{Kernel}, 62\% users determine that our agent is better on CWC dataset. This superior performance confirms the effectiveness of our approach.

\subsection{Qualitative Study}
To more intuitively demonstrate the superior performance of our agent, two conversations in user study derived from the kernel agent and our DKRN agent respectively are shown in Figure~\ref{fig:qs}. Both conversations have the same target (star) and starting utterance. From Figure~\ref{fig:qs}, we can see that our agent guides the conversation towards the target more smoothly than \textbf{Kernel}, although our agent uses one more turn to achieve the target. The complete conversations of four baseline agents and our DKRN agent are presented in the supplemental material.

\section{Conclusions}
 In this work, we move forward to develop an effective target-guided open-domain conversation system that an agent can chat with user naturally and proactively guide the conversation to an end target. Specially, 
 we propose a novel dynamic knowledge routing network (DKRN) that injects the semantic knowledge relations among candidate keywords into turn-level keyword prediction for the smooth topic transition. We also propose a simple but effective dual discourse-level target-guided strategy for a better trade-off between transition smoothness and target achievement. Additionally, we construct a new large-scale dataset for building more efficient target-guided open-domain conversation system. Extensive experiments on two datasets show the superiority of our DKRN which produce more meaningful and effective target-guided conversations.

\section*{Acknowledgements}
This work was supported by the National Natural Science Foundation of China (NSFC) under Grant No.61976233.

\bibliographystyle{aaai}
\bibliography{egbib}

\end{document}